\documentclass{article}
\usepackage{spconf,amsmath,graphicx}
\usepackage{url}
\usepackage{booktabs}

\def\corpusname{MY SCIENCE TUTOR (MyST)}
\def\virtualtutor{Marni}

\title{\corpusname{}--A Large Corpus of \\Children's Conversational Speech}
\name{Sameer S. Pradhan$^{1,2}$, Ronald A. Cole$^3$, Wayne H. Ward$^4$\thanks{The research reported here was supported by the Institute of Education Sciences, U.S. Department of Education, through Grant R305B070008, and the National Science Foundation through Grants DRL 0733323 to Boulder Language Technologies and DRL 0733322 to the University of Colorado at Boulder. The opinions expressed are those of the authors and do not represent views of the Institute, the U.S. Department of Education or the National Science Foundation.}}

\address{
  $^1${\small \tt cemantix.org}, Cambridge MA, USA\\
  $^2$Linguistic Data Consortium, University of Pennsylvania, , Philadelphia PA, USA\\
  $^3$Boulder Learning Inc., Boulder CO, USA\\
  $^4$University of Colorado at Boulder, CO, USA\\
 {\small \tt pradhan@cemantix.org}}

\begin{document}
%
\maketitle
\begin{abstract}
This article describes the MyST corpus developed as part of the My Science Tutor project--one of the largest collections of children's conversational speech comprising approximately 400 hours, spanning some 230K utterances across about 10.5K virtual tutor sessions by around 1.3K third, fourth and fifth grade students.
100K of all utterances have been transcribed thus far.
The corpus is freely available\footnote{\scriptsize \tt https://myst.cemantix.org} for non-commercial use using a creative commons license. It is also available for commercial use\footnote{\scriptsize \tt https://boulderlearning.com/resources/myst-corpus/}.
To date, ten organizations have licensed the corpus for commercial use, and approximately 40 university and other not-for-profit research groups have downloaded the corpus.
It is our hope that the corpus can be used to improve automatic speech recognition algorithms, build and evaluate conversational AI agents for education, and together help accelerate development of multimodal applications to improve children’s excitement and learning about science, and help them learn remotely.
\end{abstract}
\section{Introduction}
\label{sec:intro}

According to the 2009 National Assessment of Educational Progress (NAEP, 2009), only 34 percent of fourth-graders, 30 percent of eighth-graders, and 21 percent of twelfth-graders tested as proficient in science.
A more recent assessment, in 2019, reported a statistically significant decrease in the average score for fourth graders in science\footnote{\scriptsize \url{https://www.nationsreportcard.gov/}} compared with the most recent previous assessment, in 2015.
Thus, approximately two thirds of U.S. students are not proficient in science\footnote{This does not consider the significant impact that the educational system experienced owing to the Covid-19 pandemic.}.

This article describes a resource that was the result of a 13-year project conducted between 2007 and 2019.
The project investigated improvements in students' learning proficiency in elementary school science  using conversational multimedia virtual tutor, Marni.
The operating principles for the tutor are grounded on research from education and cognitive science where it has been shown that eliciting self-explanations plays an important role~\cite{chi-1,chi-2,chi-3,hausmann-1,hausmann-2}.
Speech, language and character animation technologies play a central role because the focus of the system is on engagement and spoken explanations by students during spoken dialog with the virtual tutor.
A series of studies conducted during this project demonstrated that students who interacted with the virtual tutor achieved substantial learning gains, equivalent to students who interacted with experienced human tutors, with moderate effect sizes~\cite{ward-1,ward-2}.
Surveys of participating teachers indicate that it is feasible to incorporate the intervention into their curriculum.
Surveys given to students indicated that over 70\% of students tutored by \virtualtutor{} were more excited about studying science in the future.

\section{The MyST corpus}
The MyST children's conversational speech corpus consists of spoken dialog between 3$^{rd}$, 4$^{th}$ and 5$^{th}$ grade students, and a virtual tutor in 8 areas of science.
It consists of 393 hours of speech collected across 1,371 students.  The collection comprises a total of 228,874 utterances across 10,496 sessions. 

\subsection{Data Collection}
As part of the study, students engaged in spoken dialog with a virtual science tutor---a lifelike computer character that produced accurate lip and tongue movement synchronized with speech produced by a voice talent.
Analyses of the spoken dialog sessions indicated that, during a dialog of about 15 minutes, tutors and students produced about the same amount of speech, around 5 minutes each.
This approach was used to develop over 100 tutorial dialog sessions, of about 15 minutes each.
\nocite{pradhan-etal-myst-corpus}
The MyST corpus was collected in two stages---Phase I and Phase II.  
In both phases, the scientific content covered is aligned to classroom science content of Full Option Science System (FOSS) modules, which typically last 8 weeks during the school year.  
FOSS is used by over 1 million children in over 100,000 classrooms in all 50 states in the U.S.  
FOSS modules are centered on science investigations.  
There are typically 4 Investigations in a module (e.g., in the Magnetism and Electricity module, the 4 investigations are Magnetism, Serial circuits, Parallel Circuits, and Electromagnetism).  
Each Investigation has 3 to 4 classroom ``investigation parts'' where groups of students work together to, for example, build a serial circuit to make a motor run, and record their observations in science notebooks.  
Shortly after conducting an ``investigation part'', students interact one-on-one with a virtual tutor for 15-20 minutes.  
The tutor asks the student questions about science presented in illustrations, animations or interactive simulations, with follow-up questions designed to stimulate reasoning and help students construct accurate explanations.

The system is \textit{strict turn-taking}; the tutor presents information, asks a question and waits for the student to respond.  
Students wear headsets with close-talking noise-canceling microphones.  
To respond, the student presses the spacebar on the laptop, holds it down while speaking, and releases it when done.  
Each student turn is recorded as a separate audio file.  
When transcribed, an utterance level transcript file is created for each audio file.  
No identifying information is stored with the data, only anonymized codes for schools and students.  
All students and their parents signed consent forms allowing us to distribute their anonymous speech data.

\subsection{Transcription}
Roughly 45\% of all utterances have been transcribed at the word level.
Phase I of the project used rich (slow, expensive) transcription guidelines\footnote{{\scriptsize \tt https://cemantix.org/myst/phase-i-guidelines/}}---the ones typically used by speech recognition researchers.
However, for the purposes of this project, that level of detail was not required in the transcriptions, and during Phase II, a reduced (quick, cheaper) version of those guidelines\footnote{{\scriptsize \tt https://cemantix.org/myst/phase-ii-guidelines/}} was used, allowing transcription of more data.

\subsection{Data Composition}
Some characteristics of the data collected in the two phases are described below.
Phase I comprised sessions from students in grades 3-5 across four science modules.
All the sessions from this phase have been transcribed using rich transcription guidelines.
Phase II comprised sessions from students in grades 4-5.
It included five modules, with an average of 10 parts each.
Table~\ref{tab:science-modules} lists the modules included in each phase.
Table~\ref{tab:cospus-size} lists the size of the corpus based on a few different parameters.

\begin{table}[th]
  \caption{Science modules covered in Phase I and II}
  \label{tab:science-modules}
  \centering
  \begin{tabular}{cll}
    \toprule
    \textbf{Phase} & \multicolumn{1}{c}{\textbf{Module}} & \multicolumn{1}{c}{\textbf{Description}} \\
    \midrule

    I  & {\bf MS}  & Mixtures and Solutions      \\
       & {\bf ME}  & Magnetism and Electricity   \\
       & {\bf VB}  & Variables                   \\
       & {\bf WA}  & Water                       \\
    \midrule
    II & {\bf EE}  & Energy and Electromagnetism \\
       & {\bf LS}  & Living Systems              \\
       & {\bf MX}  & Mixtures                    \\
       & {\bf SRL} & Soil, Rocks and Landforms   \\    
       & {\bf SMP} & Sun, Moon and Planets       \\

    \bottomrule
  \end{tabular}
\end{table}

\begin{table}[h]
  \caption{Size of corpus based on a few different criteria.}
  \label{tab:cospus-size}
  \centering
  \begin{tabular}{lr@{}rr@{}r}
    \toprule
    \multicolumn{1}{c}{\textbf{Description}} & \multicolumn{4}{c}{\textbf{Phase}} \\
\cline{2-5}
                                             & \multicolumn{2}{c}{\textbf{I}}     & \multicolumn{2}{c}{\textbf{II}} \\
\cline{2-5}
                                             & Count~ & (Hours) & Count~ & (Hours) \\

    \midrule

    Number of Students     & 421  &                 & 950   &                  \\
    Number of Sessions     & 1509 & (102)~~           & 8,987 & (291)~~            \\
    Transcribed Sessions   & 1509 & (102)~~           & 1,426 & (\hspace{2mm}95)~~ \\
    Untranscribed Sessions & 0    & (\hspace{3mm}0)~~ & 3,711 & (196)~~            \\

    \bottomrule
  \end{tabular}
\end{table}

\subsection{Corpus Structure}

The directory structure for the corpus is as shown in Figure~\ref{fig:myst-structure} below. Variables are enclosed in angle-brackets ({\scriptsize \tt <variable>}) and can take values as described immediately after.

\begin{figure}[t]
  \caption{The MyST Corpus Structure.}
  \centering
  \includegraphics[width=\linewidth]{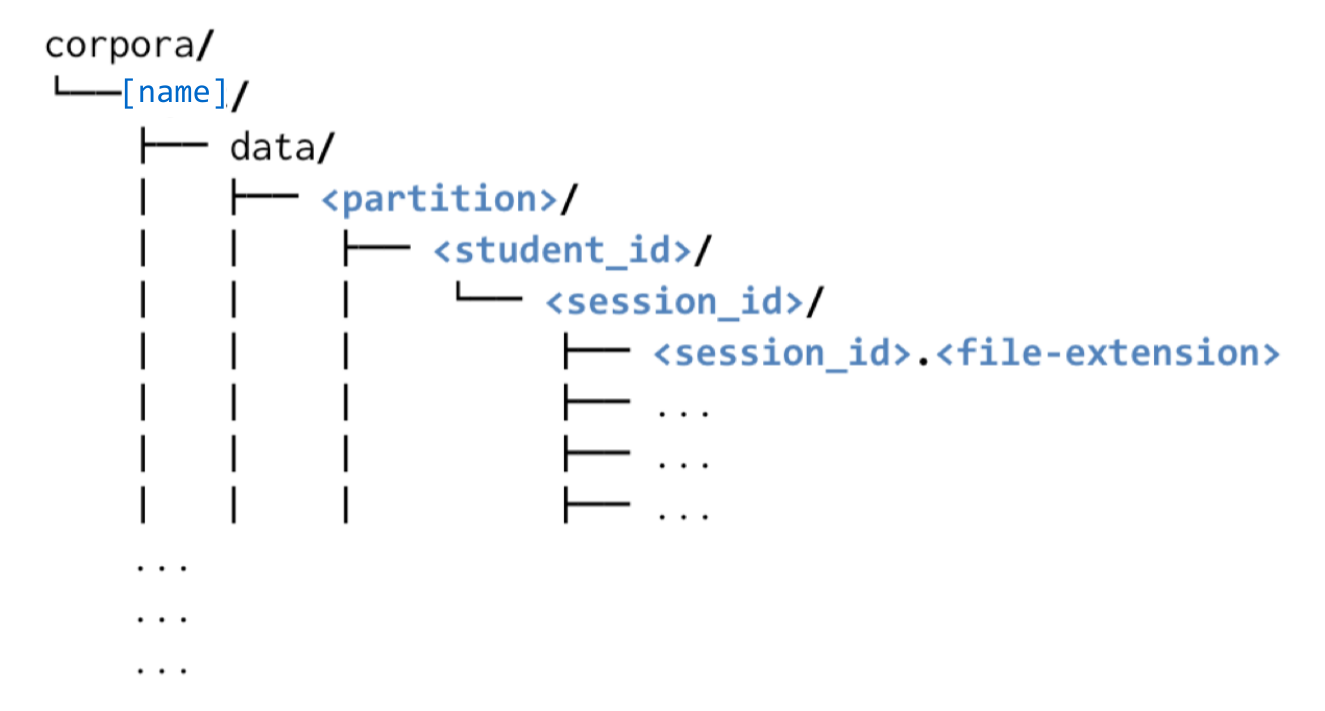}
  \label{fig:myst-structure}
\end{figure}

{\scriptsize \tt <partition>} is one of ``train'', ``development'' or ``test''.
{\scriptsize \tt <student\_id>} is a 6-digit ID with the first 3 digits representing the school code and the next 3 digits the student number.
{\scriptsize \tt <session\_id>} is the ID for a particular session and is further represented as
{\scriptsize \tt <corpus>}\_{\scriptsize \tt <student\_id>}\_{\scriptsize \tt <date>}\_{\scriptsize \tt <time>} \_{\scriptsize \tt <module>}\_{\scriptsize \tt <investigation>}.{\scriptsize \tt <part>}. {\scriptsize \tt <date>} is represented as {\scriptsize \tt <YYYY>}-{\scriptsize \tt <MM>}-{\scriptsize \tt <DD>}. {\scriptsize \tt <time>} is represented as {\scriptsize \tt <hh>}-{\scriptsize \tt <mm>}-{\scriptsize \tt <ss>}, where {\scriptsize \tt <hh>}, {\scriptsize \tt <mm>} and {\scriptsize \tt <ss>} represent two digit hour, minutes and seconds respectively\footnote{In Phase I, we did not capture hour/minute/second for each session, so the corresponding fields for sessions in Phase I are set to {\scriptsize \tt 00}}.  {\scriptsize \tt <module>} is a two- or three-character string enumerated in Table~\ref{tab:science-modules} earlier. {\scriptsize \tt <investigation>} is a decimal number representing the respective investigation for a module.  {\scriptsize \tt <part>} is the utterance ID within a session. Numbers {\scriptsize \tt 001} onward represent the index of each utterance in a session\footnote{{\tt 000} is reserved to represent the entire session.}. {\scriptsize \tt <file-extension>} is one of {\scriptsize \tt .flac} or {\scriptsize \tt .trn}. {\scriptsize \tt .flac} is the compressed audio file and {\scriptsize \tt .trn} is the transcription of the corresponding audio file.

\section{Data Cleanup and Pre-processing}
We did a pass over the corpus to clean up various types of errors that could be identified using statistics on the underlying audio and potentially erroneous data collection.

\subsection{Data Provenance}
\subsubsection{Consent}
The University of Colorado’s Institutional Review Board reviewed and approved all components of the My Science Tutor project to assure student privacy.
All utterances in the corpus were signed by a student’s parent or guardian, and by the student.
The review board approved the Parental Consent forms and the Student Assent forms.
The final Parental Consent and Student Assent forms approved by the IRB explicitly provide permission for anonymous student speech data and transcriptions to be distributed for both research and commercial use.
We manually verified that we had parental consent and student assent for every student in the corpus.

\subsection{Session Quality}
Bad—empty or corrupted sessions were removed using simple heuristics and based on missing data.

\subsubsection{Session Length}
Sessions that were less than a certain minimal threshold ($<$~10 minutes), or longer than a certain maximum threshold ($>$~1 hour) were inspected and corrected or removed.

\subsubsection{Missing audio files}
Sessions that were missing audio files for a significant number of utterances were deleted.

\subsection{Audio Quality}
All utterances were processed to identify all possible unacceptable recordings and were removed from the database. 
We performed the following checks for audio quality.

\subsubsection{Clipping Rate}
If there was a significant number of frames (exceeding a certain threshold) that were clipped, we removed or marked the audio file. 
We removed them if it impacted more than a certain fraction of utterances in a session. 
In which case we also removed the session from the release. 
If only a small number of files had large fraction of clipping, we tagged them in a report file, so that the users can determine whether to include or exclude that data from their study.

\subsubsection{Silence}
Sometimes there are significant amounts of leading and trailing silence in the audio files. 
We trimmed all such silence except for a small fraction at the beginning and end of the utterance. 
We did not, however, remove or compress silence that occurred within an utterance.

\subsubsection{Background Noise}
Utterances with a significant amount of noise or cross talk were removed. 
This was only possible for the cases that were transcribed or fell in the fraction of sample utterances that we manually verified.

\subsection{Transcription Quality}
We fixed obvious spelling errors in the transcriptions. 
We tried to retain explicitly mispronounced words as much as possible.

\subsection{Updated Pronunciation Dictionary}
We also make available an updated pronunciation dictionary. 
We used CMU’s pronunciation dictionary as a starting point and added words that were novel to this corpus.
The updated pronunciation dictionary is part of the corpus release.

\section{Evaluation}

For the convenience of the ASR community, we partitioned and structured the corpus upfront into training, development and test sets as three separate directories in the corpus release.

\subsection{Evaluation Partitions}
These partitions were generated using stratified sampling strategy thus ensuring that they reasonably represent each of the science module in MyST, proportionately represent each phase, and each student is present in only one of the three partitions. 
We also included untranscribed data in all partitions in order to be able to allow limited semi-supervised training data augmentation using the untranscribed portions of the data, with an additional advantage of pseudo-unseen data---in the form of transcriptions that are as yet absent.

\begin{table}[h]
  \caption{Experimental partitions for the MyST corpus}
  \label{tab:myst-partitions}
  \centering
  \begin{tabular}{llrrrr}
    \toprule
    \multicolumn{1}{c}{\textbf{Phase}} & \multicolumn{1}{c}{\textbf{Science}} & \multicolumn{3}{c}{\textbf{Experiment Partition}} & \multicolumn{1}{c}{\textbf{Overall}} \\
\cline{3-5}
    \multicolumn{1}{c}{\textbf{}} & \multicolumn{1}{c}{\textbf{Module}} & \multicolumn{1}{c}{\textbf{Train}}     & \multicolumn{1}{c}{\textbf{Dev.}} & \multicolumn{1}{c}{\textbf{Test}} & \\
\cline{3-5}
                               &              & (Hrs.) & (Hrs.) & (Hrs.) & (Hrs.) \\

    \midrule

I  & MS      & 31  & 5  & 5  & 41  \\
   & ME      & 30  & 4  & 4  & 38  \\
   & VB      & 14  & 2  & 2  & 18  \\
   & WA      & 4   & 1  & 1  & 6   \\
\hline
II & EE      & 114 & 16 & 14 & 144 \\
   & LS      & 75  & 4  & 4  & 83  \\
   & MX      & 29  & 5  & 7  & 41  \\
   & SRL     & 16  & 2  & 1  & 19  \\
   & SMP     & 2   & 1  & 1  & 4   \\
\hline
   & Overall & 315 & 40 & 39 & 393 \\
    \bottomrule
  \end{tabular}
\end{table}

\subsection{Experimental Setup}
We used SpeechBrain \cite{ravanelli2021speechbrain} speech toolkit for our experiments. More specifically, we used an end-to-end transformer model. We fine-tuned the model, trained on LibreSpeech model, using the MyST training set. Owing to memory limitations, we were only able to use utterances less than 30 secs. during training.  

\subsection{Word Error Rate}
We use the traditional evaluation metric of word error rate (WER) to report ASR performance.  In spite of several checks during the preparation of the corpus release, we were informed of a few new transcription errors. We corrected the errors in the test set.  For this work, we removed most of the suspicious cases from the training set and retrained the model using the filtered set of utterances. Given the amount of utterances in the training data and the very small fraction of the filtered instances, we did not see any noticeable difference in the performance on the test set. 

We plan to make another quality control pass through the corpus to correct residual errors in the development and training set and release an updated version of the corpus in the near future. We will also address other issues that arise as the corpus is used by the larger research community.

Table~\ref{tab:wer} shows the word error rate on the uncorrected and corrected version of the test set. 

\subsection{Replicability}
It is important that the ASR community report consistent and comparable WER on the MyST corpus to enable fair comparison across improved architectures. In order to facilitate that we make details of the evaluation setup and the exact configuration available at the corpus {\tt git} repository\footnote{\scriptsize \tt https://myst.cemantix.org} of the corpus. This mechanism should enable such consistent, replicable evaluations.

\begin{table}[th]
  \caption{Word Error Rate on the MyST test set for ASR Model trained only using only the training and development sets.}
  \label{tab:wer}
  \centering
  \begin{tabular}{lcc}
    \toprule

    \multicolumn{3}{c}{ \textbf{MyST Test Set}}\\ 
                       & \multicolumn{1}{c}{ Un-Corrected (\%)} & \multicolumn{1}{c}{Corrected (\%)} \\
    \midrule
    {\bf WER}          & 11.6                     & 10.0                     \\
    \hline
    {Insertions}       & 3.4                     & 2.9                     \\
    {Substitutions}    & 5.5                      & 5.1                      \\
    {Deletions}        & 2.2                      & 3.2                     \\

    \bottomrule
  \end{tabular}
\end{table}

\section{Related Work}

Over the years researchers have created several speech corpora for the analysis of children's speech. Below are a few that are typically used for ASR evaluation. For a more thorough empirical evaluation of various end-to-end ASR systems specifically focused on children's speech can be found in \cite{shivakumar2022end}.

\begin{itemize}
\item CID children’s speech corpus (American English, read speech, 436 children aged between 5 and 17 years) \cite{lee1999acoustics}

\item CMU Kid’s speech corpus (American English, read speech, 76 children, aged between 6 and 11 years) \cite{eskenazi1996kids}

\item CU Kid’s Prompted and Read Speech corpus (American English, read speech, 663 children, aged between 4 and 11 years) \cite{cole-et-al-2006},

\item CU Kid’s Read and Summarized Story corpus (American English, spontaneous speech, 326 children, aged between 6 and 11 years) \cite{cole-et-al-2006b},

\item OGI Kid’s speech corpus (English, read speech, 1100 children, aged between 5 and 15 years) \cite{shobaki2000ogi}.

\item BIRMINGHAM corpus (British English, 159 children, aged between 4 and 14 years, part of corpus PF-STAR) \cite{darcy05_interspeech}
\end{itemize}

\section{Conclusion}
Improvements in automatic transcription of children's speech--especially spontaneous conversations--can open doors to transformational applications in various areas.  Improved applications for use in education and in clinical diagnosis have the potential of making a significant global impact.  In spite of an exponentially large collection of data at our finger tips, it is difficult to get access to a reasonably large collection of specific kinds of data such as children's speech which are required by the data hungry end-to-end machine learning algorithms.  One of the larger models for reporting performance on children's speech \cite{liao2015large} used roughly 20K training utterances. However, the data underlying this study is not generally available which is a necessary requirement for open, replicable research. Our hope is that the large MyST corpus of children's conversational speech will allow researchers to improve upon consistent evaluation benchmarks.

\bibliographystyle{IEEEbib}
\bibliography{refs}

\end{document}